%% file: acl2020.tex
\documentclass[11pt,a4paper]{article}
\usepackage[]{naacl2021}
\usepackage{times}
\usepackage{latexsym}

\usepackage{microtype}

\usepackage{tabularx, booktabs}
\usepackage{multirow}
\usepackage[inline]{enumitem}
\usepackage{wrapfig}
\usepackage{array}
\usepackage{diagbox}
\usepackage{xspace}
\usepackage{url}
\usepackage{graphicx}

\usepackage[noend,ruled]{algorithm2e}
\usepackage{amsmath}
\usepackage{tikz}
\usepackage{subfigure}
 
\def\@fnsymbol#1{\ensuremath{\ifcase#1\or *\or \dagger\or \ddagger\or
   \mathsection\or \mathparagraph\or \|\or **\or \dagger\dagger
   \or \ddagger\ddagger \else\@ctrerr\fi}}
\newcommand{\ssymbol}[1]{^{\@fnsymbol{#1}}}

\newcommand{\smallsection}[1]{\noindent{\textbf{#1.}}}
\newcommand{\our}{\mbox{X-Class}\xspace}

\title{\our: Text Classification with Extremely Weak Supervision}

\author{
Zihan Wang $^1\;\;\;\;\;$ Dheeraj Mekala $^1\;\;\;\;\;$ Jingbo Shang $^{1,2}$ \\
\small $^1$ Department of Computer Science and Engineering, University of California San Diego, CA, USA \\
\small $^2$ Hal\i c\i o\u glu Data Science Institute, University of California San Diego, CA, USA \\
\small \texttt{\{ziw224, dmekala, jshang\}@ucsd.edu}
}
\date{}

\begin{document}
\maketitle
\begin{abstract}
\input{0-abstract}
\end{abstract}
\input{1-introduction}
\input{2-problem}
\input{3-method}

\input{4-experiments}
\input{5-extension}

\input{6-related}

\input{7-conclusion}

\bibliography{anthology,acl2020}
\bibliographystyle{acl_natbib}

\appendix

\end{document}

%% file: 0-abstract.tex
In this paper, we explore text classification with \textit{extremely weak supervision}, i.e., only relying on the surface text of class names.
This is a more challenging setting than the seed-driven weak supervision, which allows a few seed words per class.
We opt to attack this problem from a representation learning perspective---ideal document representations should lead to nearly the same results between clustering and the desired classification.
In particular, one can classify the same corpus differently (e.g., based on topics and locations), so document representations should be adaptive to the given class names.
We propose a novel framework \our to realize the adaptive representations.
Specifically, we first estimate class representations by incrementally adding the most similar word to each class until inconsistency arises.
Following a tailored mixture of class attention mechanisms, we obtain the document representation via a weighted average of contextualized word representations.
With the prior of each document assigned to its nearest class, we then cluster and align the documents to classes. 
Finally, we pick the most confident documents from each cluster to train a text classifier. 
Extensive experiments demonstrate that \our can rival and even outperform seed-driven weakly supervised methods on 7 benchmark datasets.

%% file: 1-introduction.tex
\section{Introduction}
Weak supervision has been recently explored in text classification to save human effort.
Typical forms of weak supervision include a few labeled documents per class~\cite{DBLP:conf/cikm/MengSZ018,DBLP:conf/emnlp/JoC19}, a few seed words per class~\cite{DBLP:conf/cikm/MengSZ018, DBLP:conf/www/MengHWWZZ020, DBLP:conf/acl/MekalaS20,DBLP:conf/emnlp/MekalaS20}, and other similar open-data~\cite{DBLP:conf/emnlp/YinHR19}.
Though much weaker than a fully annotated corpus, these forms still require non-trivial, corpus-specific knowledge from experts.
For example, nominating seed words requires experts to consider their relevance to not only the desired classes but also the input corpus;
To acquire a few labeled documents per class, unless the classes are balanced, one  needs to sample and annotate a much larger number of documents to cover the minority class. 

In this paper, we focus on \textit{extremely weak supervision}, i.e., only relying on the surface text of class names.
This setting is much more challenging than the ones above, and can be considered as almost-unsupervised text classification. 

\begin{figure}
    \centering
    \subfigure[NYT-Topics]{
        \centering
        \includegraphics[width=0.46\linewidth]{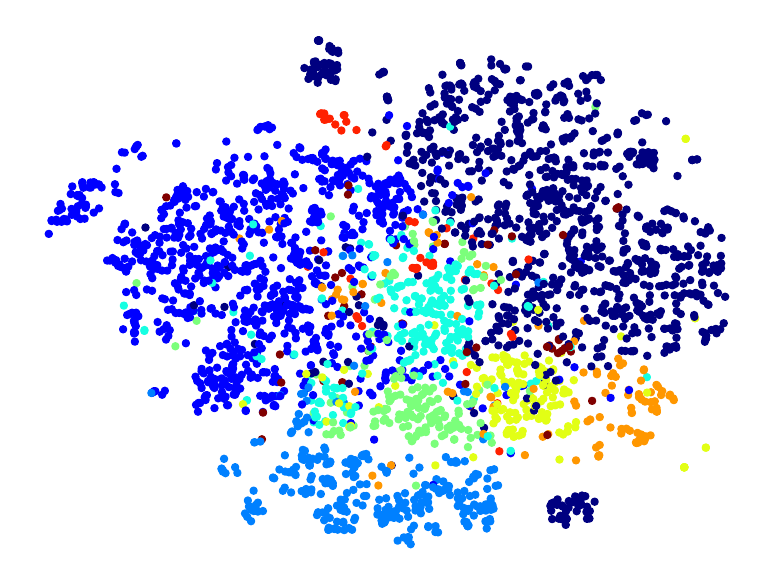}
        \vspace{-3mm}
    }
    \subfigure[NYT-Locations]{
        \centering
        \includegraphics[width=0.46\linewidth]{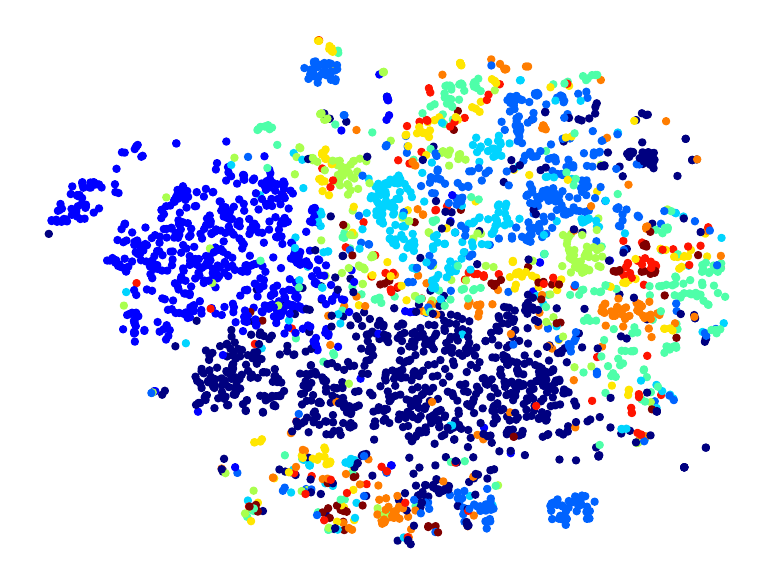}
        \vspace{-3mm}
    }
    \vspace{-3mm}
    \caption{Visualizations of the same news corpus using Average BERT Representations on two criteria. Colors denote different classes.}%
    \label{fig:multiple_sets_classes}%
    \vspace{-3mm}
\end{figure}

\begin{figure*}
    \centering
    \includegraphics[width=\linewidth]{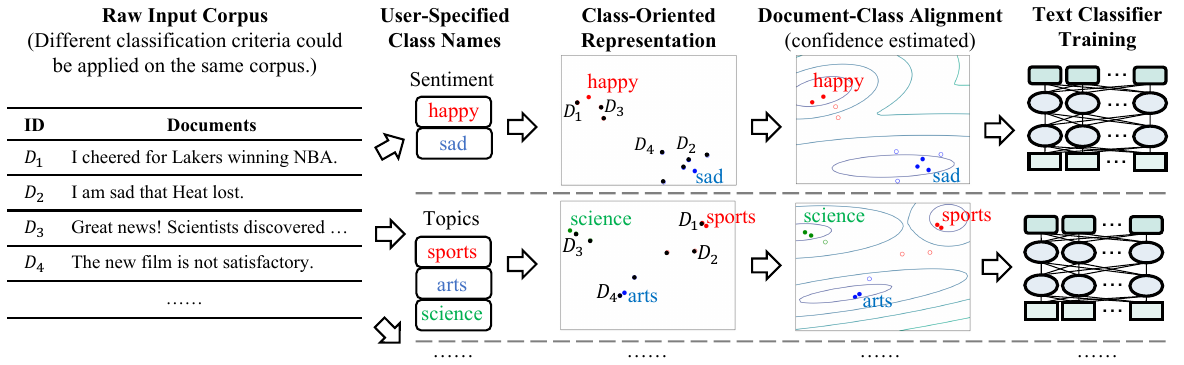}
    \caption{An overview of our \our. Given a raw input corpus and user-specified class names, we first estimate a class-oriented representation for each document. And then, we align documents to classes with confidence scores by clustering. Finally, we train a supervised model (e.g., BERT) on the confident document-class pairs.} 
    \label{fig:pipeline}
    \vspace{-3mm}
\end{figure*}

We opt to attack this problem from a representation learning perspective---ideal document representations should lead to nearly the same result between clustering and the desired classification.
Recent advances in contextualized representation learning using neural language models have demonstrated the capability of clustering text to domains with high accuracy~\cite{DBLP:conf/acl/AharoniG20}.
Specifically, a simple average of word representations is sufficient to group documents on the same topic together. 
However, the same corpus could be classified using various criteria other than topics, such as locations and sentiments.
As visualized in Figure~\ref{fig:multiple_sets_classes}, such class-invariant representations separate topics well but mix up locations.
Therefore, it is a necessity to make document representations adaptive to the user-specified class names.

We propose a novel framework \our to conduct text classification with extremely weak supervision, as illustrated in Figure~\ref{fig:pipeline}.
Firstly, we estimate class representations by incrementally adding the most similar word to each class and recalculating its representation.
Following a tailored mixture of class attention mechanisms, we obtain the document representation via a weighted average of contextualized word representations.
These representations are based on pre-trained neural language models, and they are supposed to be in the same latent space.
We then adopt clustering methods (e.g., Gaussian Mixture Models) to group the documents into $K$ clusters, where $K$ is the number of desired classes.
The clustering method is initialized with the prior knowledge of each document assigned to its nearest class.
We preserve this assignment so we can easily align the final clusters to the classes.
In the end, we pick confident documents from each cluster to form a pseudo training set, based on which, we can train any document classifier. 
In our implementation, we use BERT as both the pre-trained language model and the text classifier.
Compared with existing weakly supervised methods, \our has a stronger and more consistent performance on 7 benchmark datasets, despite some of them using at least 3 seed words per class. 
It is also worth mentioning that \our has a much more mild requirement on the existence of class names in the corpus, whereas existing methods rely on the variety of contexts of the class names.

Our contributions are summarized as follows. 
\begin{itemize}[leftmargin=*,nosep]
    \item We advocate an important but not-well-studied problem of text classification with extremely weak supervision.
    \item We develop a novel framework \our to attack this problem from a representation learning perspective. It estimates high-quality, class-oriented document representations based on pre-trained neural language models so that the confident clustering examples could form pseudo training set for any document classifiers to train on.
    \item We show that on 7 benchmark datasets, \our achieves comparable and even better performance than existing weakly supervised methods that require more human effort.
\end{itemize}

\smallsection{Reproducibility}
We will release both datasets and codes on Github\footnote{\url{https://github.com/ZihanWangKi/XClass}}.

%% file: 2-problem.tex
\section{Preliminaries}

In this section, we formally define the problem of text classification with extremely weak supervision. 
And then, we brief on some preliminaries about BERT~\cite{DBLP:conf/naacl/DevlinCLT19}, Attention~\cite{DBLP:conf/emnlp/LuongPM15} and Gaussian Mixture Models.

\smallsection{Problem Formulation}
The extremely weak supervision setting confines our input to only a set of documents $D_i, i \in \{1, ..., n\}$ and a list of class names $c_j, j \in \{1, ..., k\}$. 
The class names here are expected to provide hints about the desired classification objective, considering that different criteria (e.g., topics, sentiments, and locations) could classify the same set of documents.
Our goal is to build a classifier to categorize a (new) document into one of the classes based on the class names. 

Seed-driven weak supervision requires \emph{carefully designed} label-indicative keywords that concisely define what a class represents.
This requires human experts to understand the corpus extensively.
One of our motivations is to relax this burdensome requirement.
Interestingly, in experiments, our proposed \our using extremely weak supervision can offer comparable and even better performance than the seed-driven methods.

\smallsection{BERT}
BERT is a pre-trained masked language model with a transformer structure~\cite{DBLP:conf/naacl/DevlinCLT19}. 
It takes one or more sentences as input, breaks them up into word-pieces, and generates a contextualized representation for each word-piece. To handle long documents in BERT, we apply a sliding window technique. To retrieve representations for words, we average the representations of the word's word-pieces.
BERT has been widely adopted in a large variety of NLP tasks as a backbone.
In our work, we will utilize BERT for two purposes: (1) representations for words in the documents and (2) the supervised text classifier. 

\smallsection{Attention}
Attention mechanisms assign weights to a sequence of vectors, given a context vector~\cite{DBLP:conf/emnlp/LuongPM15}. It first estimates a hidden state $\Tilde{h}_j = K(h_j, c)$ for each vector $h_j$, where $K$ is a similarity measure and $c$ is the context vector. Then, the hidden states are transformed into a distribution via a softmax function. In our work, we use attentions to assign weights to representations, which we then average them accordingly. 

\smallsection{Gaussian Mixture Model}
Gaussian Mixture Model (GMM) is a traditional clustering algorithm~\cite{DBLP:books/lib/DudaH73}. 
It assumes that each cluster is generated through a Gaussian process. 
Given an initialization of the cluster centers and the co-variance matrix, it iteratively optimizes the point-cluster memberships and the cluster parameters following an Expectation–Maximization framework. 
Unlike K-Means, it does not restrict clusters to have a perfect ball-like shape. 
Therefore, we apply GMM to cluster our document representations.


%% file: 3-method.tex
\section{Our \our Framework}

As shown in Figure~\ref{fig:pipeline}, our \our framework contains three modules: (1) class-oriented document representation estimation, (2) document-class alignment through clustering, and (3) text classifier training based on confident labels. 


\SetAlgoSkip{}
\begin{algorithm}[t]
    \caption{Class-Oriented Document Representation Estimation} \label{alg:model_algo_embedding}
    \textbf{Input}: $n$ documents $D_i$, $k$ class names $c_j$, max number of class-indicative words $T$, and attention mechanism set $\mathcal{M}$\\
    Compute $\mathbf{t}_{i,j}$ (contextualized word rep.) \\
    Compute $\mathbf{s}_w$ for all words (Eq.~\ref{eq:static_rep}) \\
    \tcp{\textcolor{blue}{class rep. estimation}}
    \For{$l = 1 \ldots k$} {
        $\mathcal{K}_{l}$ $\leftarrow$ $\langle$ $c_l$ $\rangle$ \\
        \For{$i = 2 \ldots T$} {
            Compute $\mathbf{x}_l$ based on $\mathcal{K}_{l}$ (Eq.~\ref{eq:weighted_avg})\\
            $w = \arg\max_{w \notin \mathcal{K}_{l}} sim(\mathbf{s}_w, \mathbf{x}_l)$ \\
            Compute $\mathbf{x}'_l$ based on $\mathcal{K}_{l} \oplus \langle w \rangle$\\
            \tcp{\textcolor{blue}{consistency check}}
            \If{$\mathbf{x}'_l$ changes the words in $\mathcal{K}_{l}$} {
                \textbf{break}
            }
            \Else {
                $\mathcal{K}_{l} \leftarrow \mathcal{K}_{l} \oplus  \langle w \rangle$\\
            }
        }
    }
    \tcp{\textcolor{blue}{document rep. estimation}}
    \For{$i = 1$ ... $n$} {
        \For{\textit{attention mechanism} $m \in \mathcal{M}$} {
            Rank $D_{i,j}$ according to $m$ \\
            $r_{m, j} \leftarrow$ the rank of $D_{i,j}$ \\
        }
        Rank $D_{i,j}$ according to $\prod_m r_{m, j}$ \\
        $r_j \leftarrow$ the final rank \\
        Compute $\mathbf{E}_i$ (Eq.~\ref{eq:doc_repre})
    }
    \textbf{Return} All document representations $\mathbf{E}_i$. \\
\end{algorithm}

\subsection{Class-oriented Document Representation}\label{sec:class_oriented_document_representation}

Ideally, we wish to have some document representations such that clustering algorithms can find $k$ clusters very similar to the $k$ desired classes.

We propose to estimate the document representations and class representations based on pre-trained neural language models. Algorithm~\ref{alg:model_algo_embedding} is an overview.
In our implementation, we use BERT as an example.
For each document, we want its document representation to be similar to the class representation of its desired class.

\citet{DBLP:conf/acl/AharoniG20} demonstrated that contextualized word representations generated by BERT can preserve the domain (i.e., topic) information of documents.
Specifically, they generated document representations by averaging contextualized representations of its constituent words, and they observed these document representations to be very similar among documents belonging to the same topic.
This observation motivates us to ``classify'' documents by topics in an unsupervised way.
However, this unsupervised method may not work well on criteria other than topics. 
For example, as shown in Figure~\ref{fig:multiple_sets_classes}, such document representations work well for topics but poorly for locations.

We therefore incorporate information from the given class names and obtain \textit{class-oriented} document representations.
We break down this module into two parts, (1) class representation estimation and (2) document representation estimation.

\begin{figure}[t]
    \centering
    \includegraphics[width=1\linewidth]{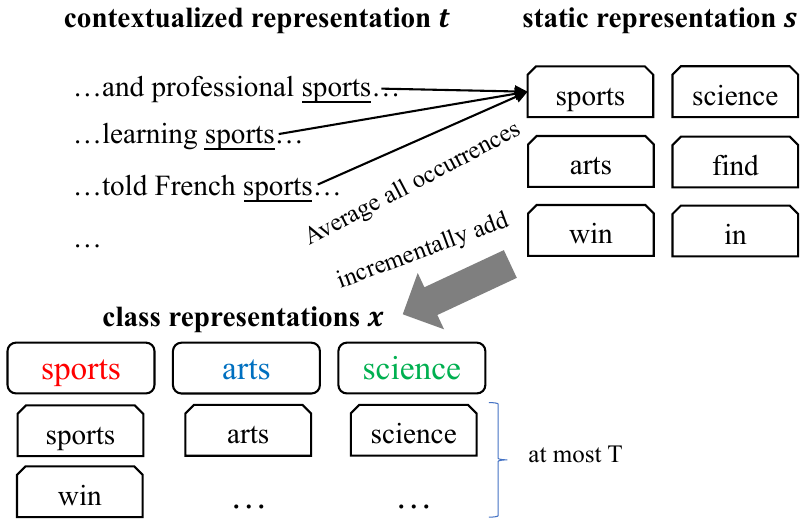}
    \vspace{-3mm}
    \caption{Overview of Our Class Rep. Estimation. 
    }%
    \label{fig:class_repre}%
    \vspace{-3mm}
\end{figure}

\smallsection{Class Representation Estimation}
    Inspired by seed-driven weakly supervised methods, we argue that a few keywords per class would be enough to understand the semantics of the user-specified classes. 
    Intuitively, the class name could be the first keyword we can start with.
    We propose to incrementally add new keywords to each class to enrich our understanding.
    
    Figure~\ref{fig:class_repre} shows an overview of our class representation estimation.
    First, for each word, we obtain its \textit{static representation} via averaging the contextualized representations of all its occurrences in the input corpus. For words that are broken into word-piece tokens, we average all the token representations as the word's representation.
    Then, we define the static representation $\mathbf{s}_w$ of a word $w$ as
    \begin{equation}
        \label{eq:static_rep}
        \mathbf{s}_w = \frac{\sum_{D_{i,j} = w} \mathbf{t}_{i, j}}{\sum_{D_{i,j} = w} 1}
    \end{equation}
    where $D_{i,j}$ is the $j$-th word in the document $D_i$ and $\mathbf{t}_{i, j}$ is its contextualized word representation. 
    \citet{DBLP:conf/emnlp/Ethayarajh19} adopted a similar strategy of estimating a static representation using BERT.
    Such static representations are used as anchors to initialize our understanding of the classes. 

    A straightforward way to enrich the class representation is to take a fixed number of words similar to the class name and average them to get a class representation.
    However, it suffers from two issues: (1) setting the same number of keywords for all classes may hurt the minority classes, and (2) a simple average may shift the semantics away from the class name itself.
    As an extreme example, when the 99\% of documents are talking about \textit{sports} and the rest 1\% are about \textit{politics}, it is not reasonable to add as many keywords as \textit{sports} to \textit{politics}---it will diverge the \textit{politics} representation.
    
    To address these two issues, we iteratively find the next keyword for each class and recalculate the class representation by a weighted average on all the keywords found.
    We stop this iterative process when the new representation is not consistent with the previous one.
    In this way, different classes will have a different number of keywords adaptively.
    Specifically, we define a comprehensive representation $\mathbf{x}_l$ for a class $l$ as a weighted average representation based on a ranked list of keywords $\mathcal{K}_l$.
    The top-ranked keywords are expected to have more similar static representations to the class representation.
    Assuming that the similarities follow Zipf's laws distribution~\cite{DBLP:conf/conll/Powers98}, we define the weight of the $i$-th keyword as $1/i$ . That is,
    \begin{equation}
        \label{eq:weighted_avg}
        \mathbf{x}_l = \frac{\sum_{i=1}^{|\mathcal{K}_l|} 1 / i \cdot \mathbf{s}_{\mathcal{K}_{l,i}}}{\sum_{i=1}^{|\mathcal{K}_l|} 1 / i}
    \end{equation}
    For a given class, the first keyword in this list is always the class name.
    In the $i$-th iteration, we retrieve the out-of-list word with the most similar static representation to the current class representation.
    We then calculate a new class representation based on all the $i+1$ words.
    We stop this expansion if we already have enough (e.g., $T = 100$) keywords, or the new class representation cannot yield the same set of top-$i$ keywords in our list.
    In our experiments, some classes indeed stop before reaching 100 keywords.
    

\begin{figure}[t]
    \centering
    \includegraphics[width=1\linewidth]{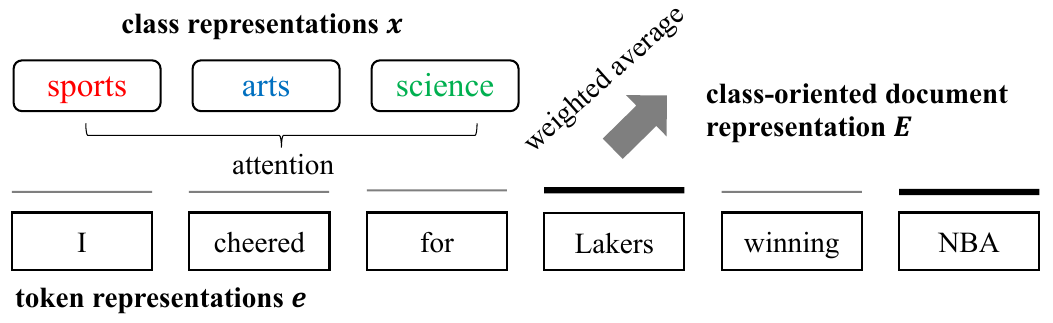}
    \vspace{-3mm}
    \caption{Overview of Our Document Rep. Estimation.
    }%
    \label{fig:document_repre}%
    \vspace{-3mm}
\end{figure}

\smallsection{Document Representation Estimation}
    Intuitively, the content of each document should stick to its underlying class. 
    For example, in the sentence ``\textit{I cheered for Lakers winning NBA}'', its content covers \textit{sports} and \textit{happy} classes, but not \textit{arts}, \textit{politics}, or \textit{sad}.
    Therefore, we assume that each word in a document is either similar to its desired class's representation or unrelated to all classes. 
    Based on this assumption, we upgrade the simple average of contextualized word representations~\cite{DBLP:conf/acl/AharoniG20} to a weighted average.
    Specifically, we follow the popular attention mechanisms to assign weights to the words based on their similarities to the class representations.
    
    Figure~\ref{fig:document_repre} shows an overview of our document representation estimation.
    We propose to employ a mixture of attention mechanisms to make it more robust.
    For the $j$-th word in the $i$-th document $D_{i,j}=w$, there are two possible representations: (1) the contextualized word representation $\mathbf{t}_{i, j}$ and (2) the static representation of this word $\mathbf{s}_w$.
    The contextualized representations disambiguate words with multiple senses by considering the context, while the static version accounts for outliers that may exist in documents.
    Therefore, it is reasonable to use either of them as the word representation $\mathbf{e}$ for attention mechanisms.
    Given the class representations $\mathbf{x}_c$, we define two attention mechanisms:
    \begin{itemize}[leftmargin=*,nosep]
        \item \textbf{one-to-one}: $h_{i,j} = \max_c \{ \mbox{cos}(\mathbf{e}, \mathbf{x_c}) \}$. It captures the maximum similarity to one class. This is useful for detecting words that are specifically similar to one class, such as \textit{NBA} to \textit{sports}. 
        \item \textbf{one-to-all}: $h_{i,j} = \mbox{cos}\left( \mathbf{e}, \mbox{avg}_c \{ \mathbf{x}_c \} \right)$ which is the similarity to the average of all classes. This ranks words by how related it is to the general set of classes in focus. 
    \end{itemize}
    Combining 2 choices of $\mathbf{e}$ and 2 choices of attention mechanisms totals 4 ways to compute each word's attention weight.
    We further fuse these attention weights in an unsupervised way.
    Instead of using the similarity values directly, we rely on the rankings.
    Specifically, we sort the words decreasingly based on attention weights to obtain 4 ranked lists.
    Following previous work~\cite{DBLP:conf/acl/MekalaS20, DBLP:conf/icdm/TaoZCJHK018}, we utilize the geometric mean of these ranks for each word and then form a unified ranked list.
    Like class representation estimation, we follow Zipf's law and assign a weight of $1/r$ to a word ranked at the $r$-th position in the end.
    Finally, we obtain the document representation $\mathbf{E}_i$ from $\mathbf{t}_{i,j}$ with these weights.
    \begin{equation}
        \label{eq:doc_repre}
        \mathbf{E}_i = \frac{\sum_j \frac{1}{r_j} \cdot \mathbf{t}_{i,j}}{\sum_j \frac{1}{j}}
    \end{equation}

\subsection{Document-Class Alignment}
\label{sec:label_document_alginment}

One straightforward idea to align the documents to classes is simply finding the most similar class based on their representations.
However, document representations not necessarily distribute ball-shape around the class representation---the dimensions in the representation can be correlated freely. 

To address this challenge, we leverage the Gaussian Mixture Model (GMM) to capture the co-variances for the clusters. Specifically, we set the number of clusters the same as the number of classes $k$ and initialize the cluster parameters based on the prior knowledge that each document $D_i$ is assigned to its nearest class $L_i$, as follows.
\begin{equation}
    L_i = \arg\max_c cos(\mathbf{E}_i, \mathbf{x}_c)
\end{equation}
We use a tied co-variance matrix across all clusters since we believe classes are similar in granularity.
We cluster the documents while remembering the class each cluster is initialized to. 
In this way, we can align the final clusters to the classes.


Considering the potential redundant noise in these representations, we also apply principal component analysis (PCA) for dimension reduction following the experience in topic clustering~\cite{DBLP:conf/acl/AharoniG20}.
By default, we fix the PCA dimension $P = 64$. 

\subsection{Text Classifier Training}
\label{sec:text_classifier_training}

The alignment between documents and classes produce high-quality pseudo labels for documents in the training set.
To generalize such knowledge to unseen text documents, we train a text classifier using these pseudo labels as ground truth.
This is a classical noisy training scenario~\cite{DBLP:journals/ml/AngluinL87,DBLP:conf/iclr/GoldbergerB17}. 
Since we know how confident we are on each instance (i.e., the posterior probability on its assigned cluster in GMM), we select the most confident ones to train a text classifier (e.g., BERT). 
By default, we set a confidence threshold $\delta = 50\%$, i.e., the top 50\% instances are selected for classifier training.

%% file: 4-experiments.tex
\input{tables/dataset_stats}

\input{tables/tbl_main}

\section{Experiments}

We conduct extensive experiments to show and ablate the performance of \our.

\subsection{Compared Methods}


We compare with two seed-driven weakly supervised methods. \textbf{WeSTClass}~\cite{DBLP:conf/cikm/MengSZ018} generates pseudo-labeled documents via word embeddings of keywords and employs a self-training module to get the final classifier. We use the CNN version of WeSTClass as it is reported to have better performance compared to the HAN version. 
\textbf{ConWea}~\cite{DBLP:conf/acl/MekalaS20} utilizes pre-trained neural language models to make the weak supervision contextualized.
In our experiments, we feed at least 3 seed words per class to these two.

We also compare with \textbf{LOTClass}~\cite{meng2020text}, which works under the extremely weak supervision setting. 
In their experiments, it mostly relies on class names but has used a few keywords to elaborate on some difficult classes. 
In our experiments, we only feed the class names to it.

We denote our method as \textbf{\our}. 
To further understand the effects of different modules, we have four ablation versions.
\textbf{\our-Rep} refers to the prior labels $L_i$ derived based on class-oriented document representation.
\textbf{\our-Align} refers to the labels obtained after document-class alignment.
\textbf{\our-ExactT} refers to not doing consistency check when estimating class representations, and having exactly T class words.
\textbf{\our-KMeans} refers to using K-Means~\cite{DBLP:journals/tit/Lloyd82} of GMM during document class alignment.

We present the performance of supervised models, serving as an upper-bound for \our.
Specifically, \textbf{Supervised} refers to a BERT model cross-validated on the training set with 2 folds (matching our confidence selection threshold). 

\subsection{Datasets}

Many different datasets have been adopted to evaluate weakly supervised methods in different works.
This makes it hard for systematic comparison.

In this paper, we pool the most popular datasets to establish a benchmark on weakly supervised text classification.
Table~\ref{tbl:dataset_stats} provides an overview of our carefully selected 7 datasets, covering different text sources (e.g., news, reviews, and Wikipedia articles) and different criteria of classes (e.g., topics, locations, and sentiment).
\begin{itemize}[leftmargin=*,nosep]
    \item \textbf{AGNews} from~\cite{DBLP:conf/nips/ZhangZL15} (used in WeSTClass and LOTClass) is for topic categorization in news from AG’s corpus.
    \item \textbf{20News} from~\cite{DBLP:conf/icml/Lang95}\footnote{\url{http://qwone.com/~jason/20Newsgroups/}} (used in WeSTClass and ConWea) is for topic categorization in news.
    \item \textbf{NYT-Small} (used in WeSTClass and ConWea) is for classifying topic in New York Times news.
    \item \textbf{NYT-Topic} (used in~\cite{DBLP:conf/www/MengHWWZZ020}) is another larger dataset collected from New York Times for topic categorization. 
    \item \textbf{NYT-Location} (used in~\cite{DBLP:conf/www/MengHWWZZ020}) is the same corpus as NYT-Topic but for locations. It is noteworthy to point out that many documents from this dataset talk about several countries simultaneously, so simply checking the location names will not lead to satisfactory results.
    \item \textbf{Yelp} from~\cite{DBLP:conf/nips/ZhangZL15} (used in WeSTClass) is for sentiment analysis in reviews.
    \item \textbf{DBpedia} from~\cite{DBLP:conf/nips/ZhangZL15} (used in LOTClass)  is for topic classification based on titles and descriptions in DBpedia.
\end{itemize}

\begin{figure*}[t]
    \centering
    \subfigure[Our Class-Oriented Document Representations]{
        \centering
        \includegraphics[width=0.15\linewidth]{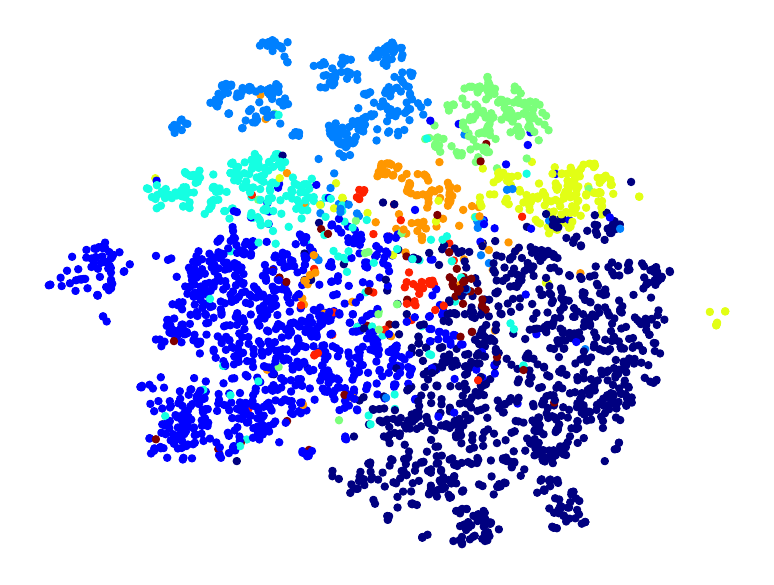}
        \includegraphics[width=0.15\linewidth]{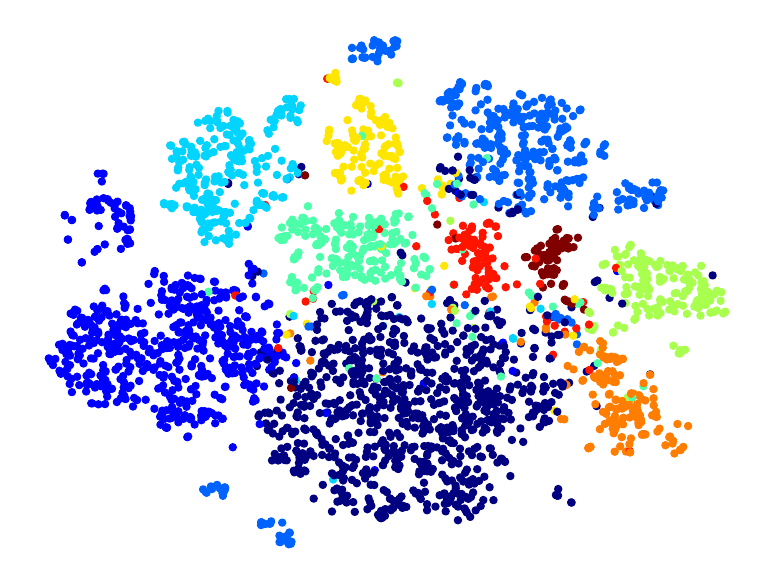}
        \includegraphics[width=0.15\linewidth]{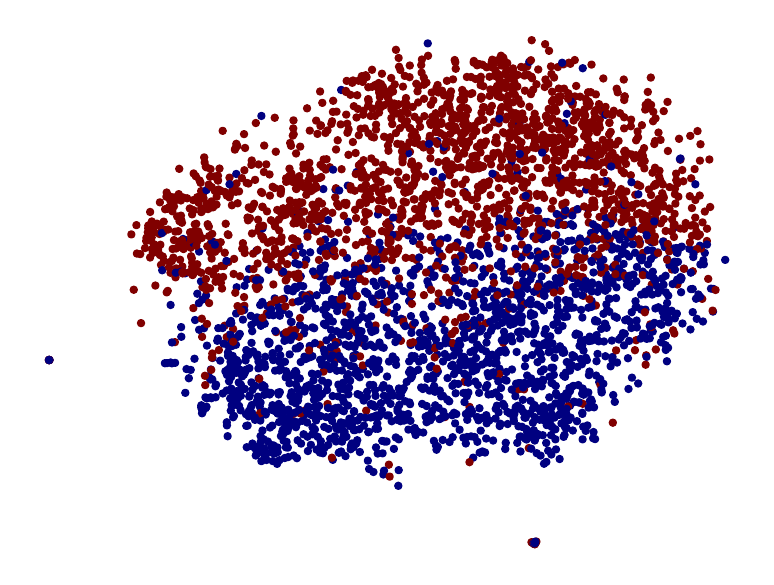}
    }
    \subfigure[Simple Average of BERT Representations]{
        \centering
        \includegraphics[width=0.15\linewidth]{figures/nyt-topics-noweight.pdf}
        \includegraphics[width=0.15\linewidth]{figures/nyt-locations-noweight.pdf}
        \includegraphics[width=0.15\linewidth]{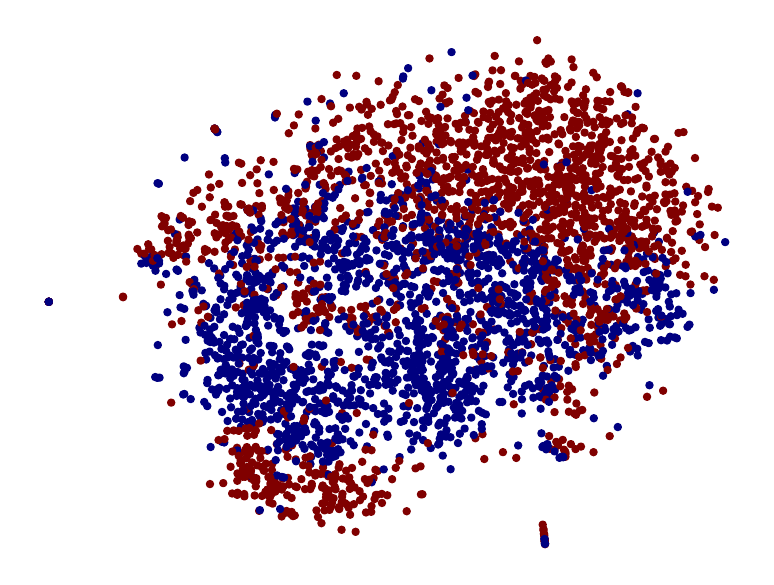}
    }
    \vspace{-3mm}
    \caption{t-SNE Visualizations of Representations. From left to right: NYT-Topics, NYT-Locations, Yelp.}
    \label{fig:weight}
    \vspace{-3mm}
\end{figure*}

\subsection{Experimental Settings}
For all \our experiments, we report the performance under one fixed random seed. 
By default, we set $T = 100, P = 64, \delta = 50\%$. 
For contextualized token representations $\mathbf{t}_{i,j}$, we use the \texttt{BERT-base-uncased} to group more occurrences of the same word.
For supervised model training, we follow BERT fine-tuning~\cite{DBLP:journals/corr/abs-1910-03771} with all hyper-parameters unchanged.

\input{tables/tbl_example_seedwords}
For both WeSTClass and ConWea, we have tried our best to find keywords for the new datasets. Table~\ref{tbl:example} shows an example on the seed words selected for them on the NYT-Small dataset. 
For LOTClass, we tune their hyper-parameters \textit{match\_threshold} and \textit{mcp\_epoch}, and report the best performance during their self-train process.


\subsection{Performance Comparison and Analysis}

From Table~\ref{tbl:main}, one can see that \our achieves the best overall performance.
It is only 1\% to 2\% away from LOTClass and ConWea on AGNews and NYT-Topics, respectively.
Note that, ConWea consumes at least 3 keywords per class.

It is noteworthy that \our can approach the supervised upper bound to a small spread, especially on the NYT-Small dataset.

\smallsection{Ablation on Modules}
\our-Rep has achieved high scores (e.g., on both NYT-Topics and NYT-Locations) showing success of our class-oriented representations.
The improvement of \our-Align over \our-Rep demonstrates the usefulness of our clustering module. 
It is also clear that the classifier training is beneficial by comparing \our and \our-Align.

\smallsection{Ablation on Consistency Check}
The consistency check in class representation estimation allows an adaptive number of keywords for each class. Without it leads to a diverged class understanding and degrading performance, as shown in Table~\ref{tbl:main}.


\smallsection{Ablation on Clustering Methods}
Table~\ref{tbl:main} also shows that K-Means performs poorly on most datasets. 
This matches our previous analysis as K-Means assumes a hard spherical boundary, while GMM models the boundary softly like an ellipse.




\subsection{Effect of Attention}

In Figure~\ref{fig:weight}, we visualize our class-oriented document representations and the unweighted variants using t-SNE~\cite{DBLP:conf/vissym/RauberFT16}. We can see that while the simple-average representations are well-separated like class-oriented representations in NYT-Topics, they are much mixed up in NYT-Locations and Yelp.
We conjecture that this is because BERT representations has topic information as its most significant feature. 

\begin{figure}[t]
    \centering
    \includegraphics[width=\linewidth]{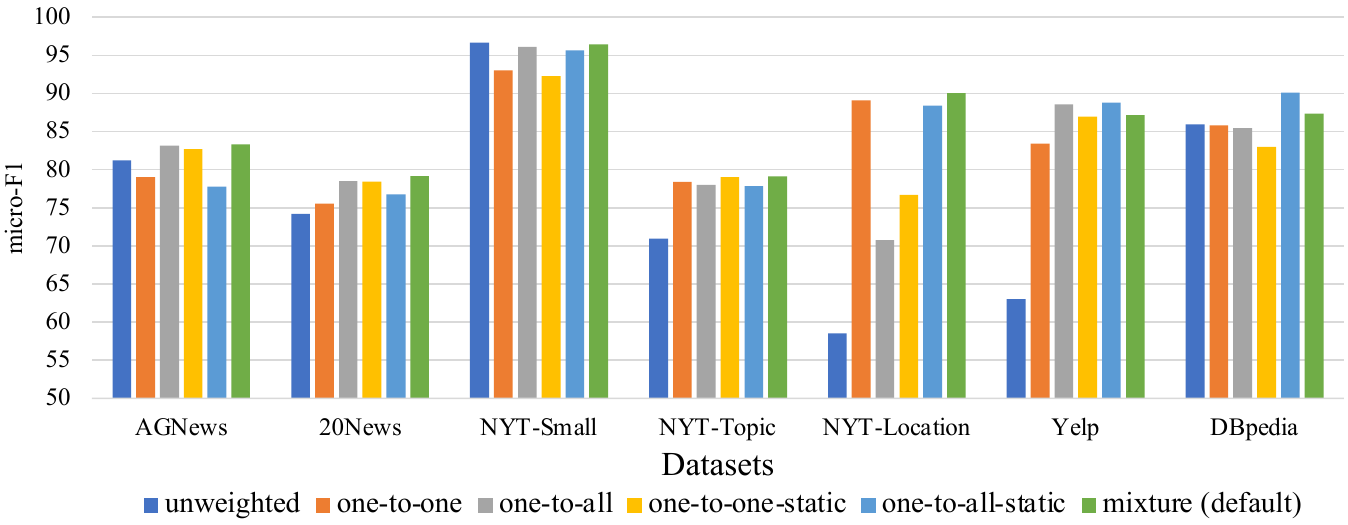}
    \vspace{-5mm}
    \caption{Effects of Attention Mechanisms. We focus on \our-Align to show their direct effects. 
    }
    \vspace{-3mm}
    \label{fig:attention_metric}
\end{figure}

We have also tried using different attention mechanisms in \our. 
From the results in Figure~\ref{fig:attention_metric}, one can see that using a single mechanism, though not under-performing much, is less stable than our proposed mixture. 
The unweighted case works well on all four datasets that focus on news topics but not good enough on locations and sentiments.


\begin{figure*}
    \centering
    \includegraphics[width=0.93\linewidth]{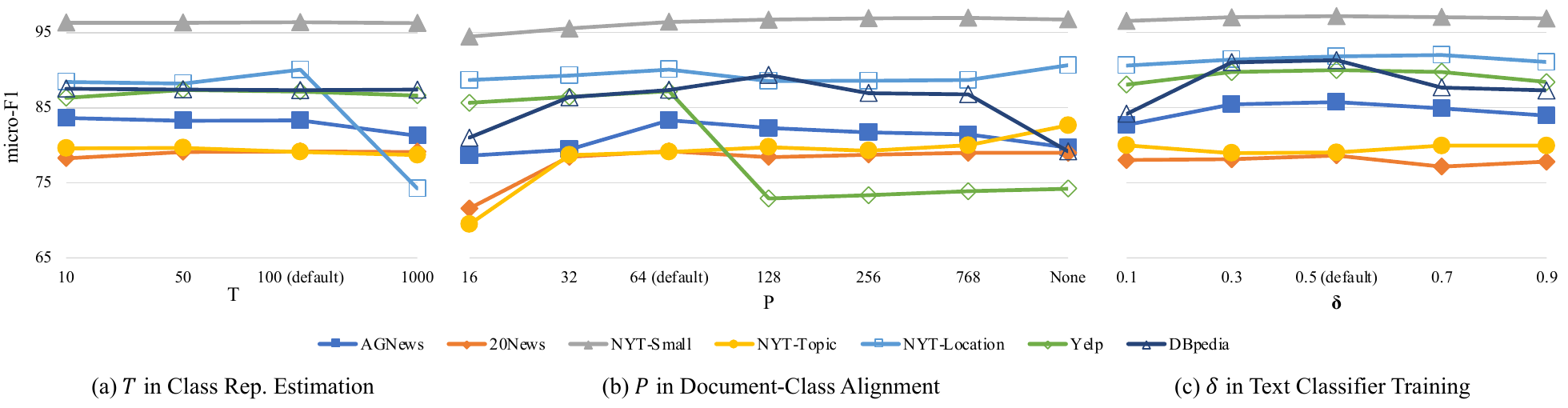}
    \vspace{-3mm}
    \caption{Hyper-parameter Sensitivity in \our. For $T$ and $P$, we report the performance of \our-Align to explore their direct effects. }
    \label{fig:hyper}
    \vspace{-3mm}
\end{figure*}

\subsection{Hyper-parameter Sensitivity in \our}

Figure~\ref{fig:hyper} visualizes the performance trend w.r.t. to the three hyper-parameters in \our, i.e., the limit of class words $T$ in class representation estimation, the PCA dimension $P$ in document-class alignment, and the confidence threshold $\delta$ in text classifier training. 

Intuitively, a class doesn't have too many highly relevant keywords. 
One can confirm this in Figure~\ref{fig:hyper}(a) as the performance of \our is relatively stable unless $T$ goes too large to 1000. 

Choosing a proper PCA dimension could prune out redundant information in the embeddings and improve the running time. 
However, if $P$ is too small or too large, it may hurt due to information loss or redundancy. 
One can observe this expected trend in Figure~\ref{fig:hyper}(b) on all datasets. 

Typically, we want to select a reasonable number of confident training samples for the text classifier training.
Too few training samples (i.e., too large $\delta$) would lead to insufficient training data.
Too many training samples (i.e., too small $\delta$) would lead to too noisy training data.
Figure~\ref{fig:hyper}(c) shows that $\delta \in [0.3, 0.9]$ is a good choice on all datasets.

%% file: tables/dataset_stats.tex
\begin{table*}[t]
\centering
\caption{An overview of our 7 benchmark datasets. They cover various domains and classification criteria. The imbalance factor of a dataset refers to the ratio of its largest class's size to the smallest class's size.}
\label{tbl:dataset_stats}
\vspace{-3mm}
\small
\begin{tabular}{l c c c c c c c} 
\toprule
\textbf{} & \textbf{AGNews} & \textbf{20News} & \textbf{NYT-Small} & \textbf{NYT-Topic} & \textbf{NYT-Location} & \textbf{Yelp} & \textbf{DBpedia}  \\
\midrule
Corpus Domain & News & News & News & News & News & Reviews & Wikipedia \\
Class Criterion & Topics & Topics & Topics & Topics & Locations & Sentiment & Ontology\\
\# of Classes & 4 & 5 & 5 & 9 & 10 & 2 & 14 \\
\# of Documents & 120,000 & 17,871 & 13,081 & 31,997 & 31,997 & 38,000 & 560,000 \\
Imbalance & 1.0 & 2.02 & 16.65 & 27.09 & 15.84 & 1.0 & 1.0 \\
\bottomrule
\end{tabular}
\end{table*}

%% file: tables/tbl_main.tex
\begin{table*}[t]
\centering
\caption{Evaluations of Compared Methods and \our. Both micro-/macro-F$_1$ scores are reported.
Supervised provides an upper bound. 
$\ssymbol{2}$ indicates the use of at least 3 seed words per class. 
$\ssymbol{3}$ indicates the use of only class names. 
$\ssymbol{4}$ refers to number coming from other papers.
ConWea is too slow on DBpedia, therefore not reported.
}
\vspace{-3mm}
\label{tbl:main}
\scalebox{0.95}{
\small
\begin{tabular}{l c c c c c c c} 
\toprule
\textbf{Model} & \textbf{AGNews} & \textbf{20News} & \textbf{NYT-Small} & \textbf{NYT-Topic} & \textbf{NYT-Location} & \textbf{Yelp} & \textbf{DBpedia}  \\
\midrule
Supervised & 93.99/93.99 & 96.45/96.42 & 97.95/95.46 & 94.29/89.90 & 95.99/94.99 & 95.7/95.7  & 98.96/98.96 \\
\midrule
WeSTClass$\ssymbol{2}$ & 82.3/82.1$\ssymbol{4}$ &  71.28/69.90 & 91.2/83.7$\ssymbol{4}$ & 68.26/57.02 & 63.15/53.22 & 81.6/81.6$\ssymbol{4}$ & 81.42/81.19\\
ConWea$\ssymbol{2}$ & 74.6/74.2 & 75.73/73.26 & 95.23/90.79 & \textbf{81.67}/\textbf{71.54} & 85.31/83.81 & 71.4/71.2 & N/A \\
LOTClass$\ssymbol{3}$ & \textbf{86.89}/\textbf{86.82} & 73.78/72.53 & 78.12/56.05 & 67.11/43.58 & 58.49/58.96 & 87.75/87.68 & 86.66/85.98 \\
\our$\ssymbol{3}$ & 85.74/85.66 & \textbf{78.62}/\textbf{77.76} & \textbf{97.18}/\textbf{94.02} & 79.02/68.55 & \textbf{91.8}/\textbf{91.98} & \textbf{90.0}/\textbf{90.0} & \textbf{91.32}/\textbf{91.17}\\
\midrule
\multicolumn{8}{c}{Ablations} \\
\midrule
\our-Rep$\ssymbol{3}$ & 77.86/76.84 & 75.37/73.7 & 92.13/83.69 & 77.06/65.05 & 86.36/88.1 & 78.0/77.19 & 74.05/71.74\\
\our-Align$\ssymbol{3}$ & 83.32/83.28 & 79.19/78.46 & 96.42/92.32 & 79.12/67.76 & 90.09/90.63 & 87.19/87.13 & 87.36/87.27\\


\our-ExactT$\ssymbol{3}$ & 84.85/84.76 & 73.95/74.13 & 97.18/94.02 & 79.18/68.96 & 88.94/88.02 & 90.0/90.0 & 88.48/88.37 \\
\our-KMeans$\ssymbol{3}$ & 81.29/81.08 & 70.79/71.18 & 94.96/89.66 & 72.83/64.79 & 93.88/92.94 & 80.6/80.56 & 65.76/66.94 \\
\bottomrule
\end{tabular}
}
\vspace{-3mm}
\end{table*}

%% file: tables/tbl_example_seedwords.tex
\begin{table}[t]
\centering
\caption{Example seed words and class names for methods for NYT-Small.}
\vspace{-3mm}
\small
\begin{tabular}{l l c} 
\toprule
\textbf{class} & Seed words & Class name for \our \\
\midrule
\multirow{2}{*}{arts} & dance,art, & \multirow{2}{*}{arts} \\
& ballet,museum & \\
\multirow{2}{*}{business} & shares,stocks, & \multirow{2}{*}{business} \\
& markets,trading & \\
\bottomrule 
\end{tabular}
\vspace{-3mm}
\label{tbl:example}
\end{table}

%% file: 5-extension.tex
\subsection{Requirements on Class Names}
\label{sec:one_occ}

Compared with previous works~\cite{DBLP:conf/cikm/MengSZ018, DBLP:conf/acl/MekalaS20, meng2020text}, our \our has a significantly more mild requirement on human-provided class names in terms of quantity and quality. We have conducted an experiment in Table~\ref{tbl:one_occ} for \our on 20News and NYT-Small by deleting all but one occurrence of a class name from the input corpus. In other words, the user-provided class name only appears once in the corpus. Interestingly, the performance of \our only drops less than 1\%, still outperforming all compared methods. In contrast, the most recent work, LOTClass~\cite{meng2020text}, requires a wide variety of contexts of class names from the input corpus to ensure the quality of generated class vocabulary in its very first step. 
\input{tables/tbl_one_occ}



\section{\our for Hierarchical Classification}

There are two straightforward ways to extend \our for hierarchical classification
(1) \textbf{\our-End}: We can give all fine-grained class names as input to \our and conduct classification in an end-to-end manner;
and
(2) \textbf{\our-Hier}: We can first give only coarse-grained class names to \our and obtain coarse-grained predictions. Then, for each coarse-grained class and its predicted documents, we further create a new \our classifier based on the fine-grained class names.

We experiment with hierarchical classification on the NYT-Small dataset, which has annotations for 26 fine-grained classes. 
We also introduce \textbf{WeSHClass}~\cite{DBLP:conf/aaai/MengSZH19}, the hierarchical version of WeSTClass, for comparison.
LOTClass is not investigated here due to its poor coarse-grained performance on this dataset.
The results in Table~\ref{tbl:finegrained} show that \our-Hier performs the best, and it is a better solution than \our-End.
We conjecture that this is because the fine-grained classes' similarities are drastically different (a pair of fine-grained classes can much similar than another pair). 
Overall, we show that we can apply our method to a hierarchy of classes.

\input{tables/tbl_finegrained}

%% file: tables/tbl_one_occ.tex
\begin{table}[t]
\centering
\renewcommand\tabcolsep{2pt}
\renewcommand\arraystretch{0.9}
\caption{Macro-F$_1$ score changes of methods when removing all but one occurrence of a class name.}
\vspace{-3mm}
\small
\begin{tabular}{l c c c c} 
\toprule
\textbf{Model} & \multicolumn{2}{c}{20News} & \multicolumn{2}{c}{NYT-Small} \\
 & Original & Removed & Original & Removed \\
\midrule


\our & 77.76 & 74.48 & 94.02 & 93.29 \\
LOTClass  & 72.53 & 8.82 & 56.05 & 29.53 \\
\bottomrule 
\end{tabular}
\vspace{-3mm}
\label{tbl:one_occ}
\end{table}

%% file: tables/tbl_finegrained.tex
\begin{table}[t]
\centering
\caption{Micro-/Macro-F$_1$ scores for Fine-grained Classification on NYT-Small. All compared methods use 3 keywords per class. LOTClass failed to discover documents with category indicative terms, thus not reported. $\ssymbol{4}$ refers to numbers coming from other papers.}

\vspace{-3mm}
\small
\begin{tabular}{l c c} 
\toprule
\textbf{Model} & \textbf{Coarse (5 classes)} & \textbf{Fine (26 classes)} \\
\midrule
WeSTClass & \multirow{ 2}{*}{91/84$\ssymbol{4}$} & 50/36$\ssymbol{4}$ \\
WeSHClass & & 87.4/63.2$\ssymbol{4}$ \\
ConWea & 95.23/90.79 & 91/79$\ssymbol{4}$ \\
\midrule
\our-End & \multirow{ 2}{*}{96.67/92.98} & 86.07/75.30\\
\our-Hier &  & 92.66/80.92 \\
\bottomrule
\end{tabular}
\vspace{-3mm}
\label{tbl:finegrained}
\end{table}

%% file: 6-related.tex
\section{Related Work}
We discuss related work from two angles.

\smallsection{Weakly supervised text classification}
Weakly supervised text classification has attracted much attention from researchers~\cite{DBLP:conf/icdm/TaoZCJHK018, DBLP:conf/www/MengHWWZZ020, DBLP:conf/acl/MekalaS20, meng2020text}. The general pipeline is to generate a set of document-class pairs to train a supervised model above them. Most previous work utilizes keywords to find such pseudo data for training, which requires an expert that understands the corpus well. In this paper, we show that it is possible to reach a similar, and often better, performance on various datasets without such guidance from experts. 

A recent work~\cite{meng2020text} also studied the same topic --- extremely weak supervision on text classification. It follows a similar idea of~\cite{DBLP:conf/www/MengHWWZZ020} and further utilizes BERT to query replacements for class names to find keywords for classes, identifying potential classes for documents via string matching. Compared with LoTClass, our \our has a less strict requirement of class names being existent in the corpus, and can work well even when there is only one occurrence (refer to Section~\ref{sec:one_occ}). 

\smallsection{BERT for topic clustering}
\citet{DBLP:conf/acl/AharoniG20} showed that document representations obtained by an average of token representations from BERT preserve domain information well. We borrow this idea to improve our document representations through clustering. Our work differs from theirs in that our document representations are guided by the given class names.

%% file: 7-conclusion.tex
\section{Conclusions and Future Work}
We propose our method \our for extremely weak supervision on text classification, which is to classify text with only class names as supervision.
\our leverages BERT representations to generate class-oriented document presentations, which we then cluster to form document-class pairs, and in the end, fed to a supervised model to train on.
We further set up benchmark datasets for this task that covers different data (news and reviews) and various class types (topics, locations, and sentiments).
Through extensive experiments, we show the strong performance and stability of our method. 

There are two directions that are possible to explore. First, focusing on the extremely weak supervision setting, we can extend to many other natural language tasks to eliminate human effort, such as Named Entity Recognition and Entity Linking. Second, based on the results on extremely weak supervision, we can expect an unsupervised version of text classification, where machines suggest class names and classify documents automatically. 

\section{Acknowledgements}
We thank all reviewers for their constructive comments; Yu Meng for valuable
discussions and comments. Our work is supported in part by NSF Convergence Accelerator under award OIA-2040727. Any opinions, findings, and conclusions or recommendations expressed herein are those of the authors and should not be interpreted as necessarily representing the views, either expressed or implied, of the U.S. Government. The U.S. Government is authorized to reproduce and distribute reprints for government purposes notwithstanding any copyright annotation hereon.


\section{Ethical Considerations}
We do not anticipate any significant ethical concerns; Text Classification is a fundamental problem in Natural Language Processing. The intended use of this work would be to classify documents, such as news articles, efficiently. A minor consideration is the potential for certain types of hidden biases to be introduced into our results, such as a biased selection of class names or language model pre-trained on biased data. We did not observe this kind of issue in our experiments, and indeed these considerations seem low-risk for the specific datasets studied here.